\documentclass[11pt,a4paper]{article}
\usepackage[hyperref]{acl2020}
\usepackage{times}
\usepackage{latexsym}
\usepackage{graphicx}
\usepackage{float}
\usepackage{amsmath}
\usepackage{paralist}

\usepackage{hyperref}
\usepackage{tablefootnote}
\usepackage{tabularx}
\usepackage{microtype}

\aclfinalcopy 
\def\aclpaperid{22} 

\title{Latent Alignment of Procedural Concepts in Multimodal Recipes}

\author{
Hossein Rajaby Faghihi,
Roshanak Mirzaee,
Sudarshan Paliwal\and
Parisa Kordjamshidi\\
Michigan State University\\
\{rajabyfa, mirzaeem, paliwal, kordjams\}@msu.edu
}
\date{}

\begin{document}
\maketitle
\begin{abstract}
We propose a novel alignment mechanism to deal with procedural reasoning on a newly released multimodal QA dataset, named RecipeQA. Our model is solving the textual cloze task which is a reading comprehension on a recipe containing images and instructions. We exploit the power of attention networks, cross-modal representations, and a latent alignment space between instructions and candidate answers to solve the problem. We introduce constrained max-pooling which refines the max-pooling operation on the alignment matrix to impose disjoint constraints among the outputs of the model. Our evaluation result indicates a 19\% improvement over the baselines. 
\end{abstract}

\section{Introduction}

Procedural reasoning by following several steps to achieve a goal is an essential part of our daily tasks. However, this is challenging for machines due to the complexity of instructions and commonsense reasoning required for understanding the procedure~\cite{mishra2018tracking,yagcioglu2018recipeqa,bosselut2017simulating}.

In this paper, we tackle the task of procedural reasoning in a multimodal setting for understanding cooking recipes.
The RecipeQA dataset~\cite{yagcioglu2018recipeqa} contains recipes from internet users. Thus, understanding the text is challenging due to the different language usage and informal nature of user-generated texts. The recipes are along with images provided by users which are taken in an unconstrained environment. This exposes a level of difficulty similar to real-world problems. 

\begin{figure*}[t]
    \centering
    \includegraphics[width=\linewidth]{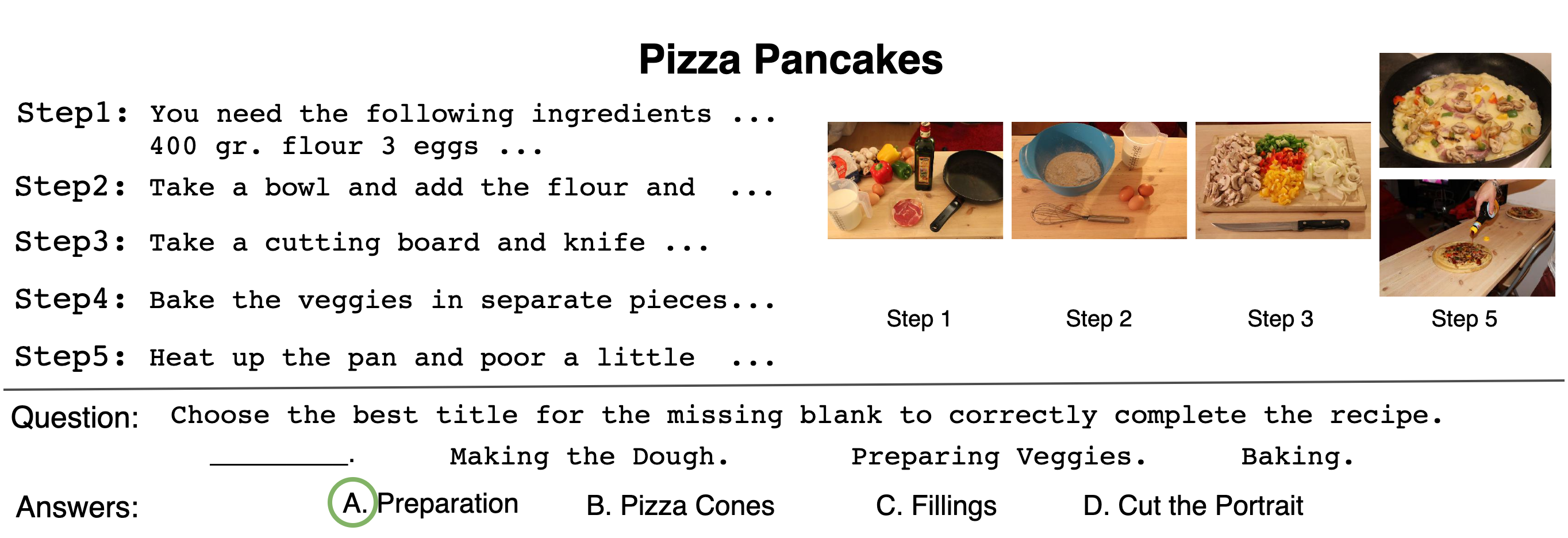}
    \caption{A sample of textual cloze task}
    \label{fig:sample}
\end{figure*}

The tasks proposed with the dataset include textual cloze, visual cloze, visual ordering, and visual coherence. 
Here, we focus on textual cloze. An example of this task is shown in Figure \ref{fig:sample}. The input to the task is a set of multimodal instructions, three textual items from the question and a placeholder to be filled by the answer. The answer has to be chosen from four options. The three question items and the correct answer make a sequence which correctly describes the steps of the recipe.

To design our model, we rely on the intuition that given question items, each answer describes exactly one step of the recipe.
Hence, we design a model to make explicit alignments between the candidate answers and each step and use those alignment results, given the question information. This alignment space is latent due to not having any direct supervision based on provided annotations. 

Using multimodal information and representations by making a joint space for comparison has been broadly investigated in the recent research ~\cite{hessel-lee-mimno-2019unsupervised,UnifiedVisual,li2019visualbert,su2020vl-bert,yu2019multimodal,fan2018stacked,tan-bansal-2019-lxmert,nam2017dual}. Our work differs from those as we do not have direct supervision on multimodal alignments. Moreover, the task we are solving uses  the sequential nature of visual and textual modality as a weak source of supervision to build a neural model to compare the textual representation of context and the answers for a given question representation.

Procedural reasoning has been investigated on different tasks~\cite{amac-etal-2019-procedural,CRCN}. While PRN~\cite{amac-etal-2019-procedural} is proposed on RecipeQA, their model does not apply to the textual cloze task. \cite{CRCN} is using procedural reasoning on multimodal information to generate a story from a sequence of images. However, the textual cloze task is about filling a blank in a sequence given a set of textual options.

Our model exploits the latent alignment space and the positional encoding of questions and answers while applying a novel approach for constraining the output space of the latent alignment. Moreover, we exploit cross-modality representations based on cross attention to investigate the benefits from information flow between images and instructions. We compare our results to the provided baselines in~\cite{yagcioglu2018recipeqa} and achieve the state-of-the-art by improving over 19\%. 



\section{Proposed Model}
We design a model to solve a structured output prediction on the textual cloze task. The intuition of our model is that the correct answer option should describe precisely one instruction, and this instruction should not be already described with  other items in the question. Hence, our model assumes the instruction and question as the context and candidate answers as an additional input to the alignment process. 
Moreover, to incorporate the order of the sequence in question items and the placeholder, we utilize a one-hot encoding vector of positions to be concatenated with the candidate answers and question items' representations.

\begin{figure*}[t]
    \centering
    \includegraphics[width=\linewidth]{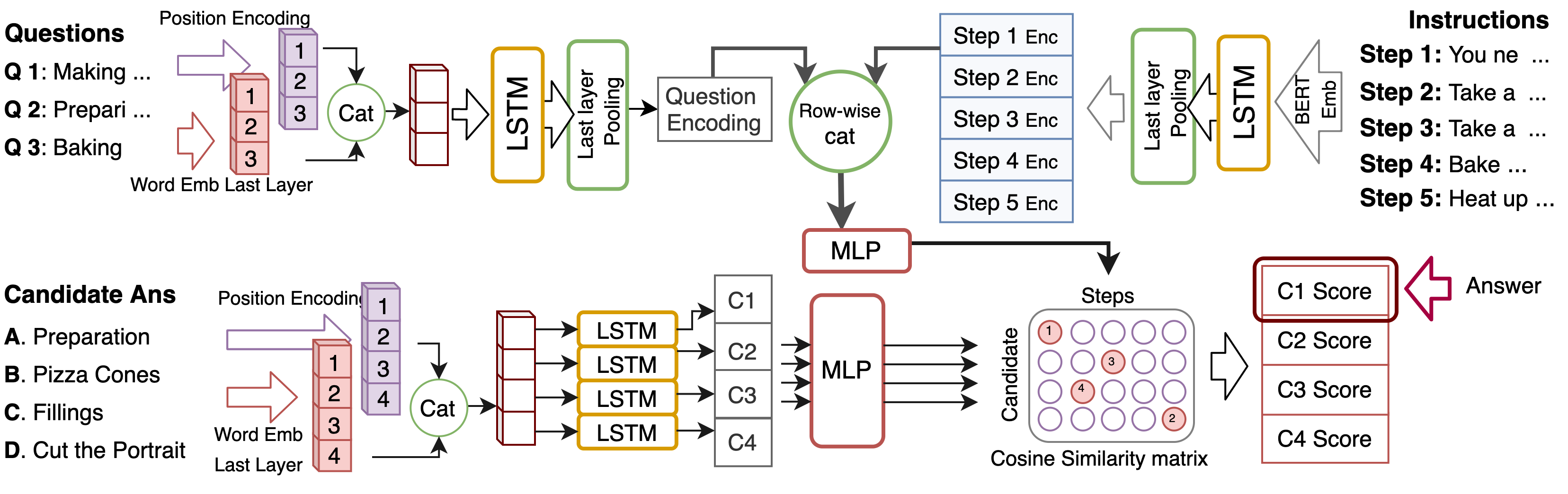}
    \caption{An overview of proposed model}
    \label{fig:model}
\end{figure*}

We give the instructions to a sentence splitter using Stanford Core NLP library~\cite{manning2014stanford}. The output is then tokenized by Flair data structure~\cite{akbik2018coling} and embedded with BERT~\cite{devlin-etal-2019-bert}. The words' embeddings are passed to an LSTM layer and the last layer is used as the instruction representation. We propose two different approaches to include images representations. These proposals are described in Section \ref{sec:ablation}. An overview of our approach is shown in Figure \ref{fig:model}.

Question representation is the last layer of an LSTM on question items. The representation of each question item is the concatenated vector of a one-hot position encoding and word embedding obtained from BERT. The candidate answers' representations are computed using the same approach. We concatenate the question representation to each instruction. Then, the similarity of each candidate answer and instruction is computed using the cosine similarity and form a similarity matrix. We use $S$ to denote the similarity matrix. The rows of this matrix are candidate answers and the columns represent the recipe steps. The value of $S_{ij}$ indicates the similarity score of candidate $i$ and step $j$.

\begin{figure}[!h]
    \centering
    \includegraphics[width=\linewidth]{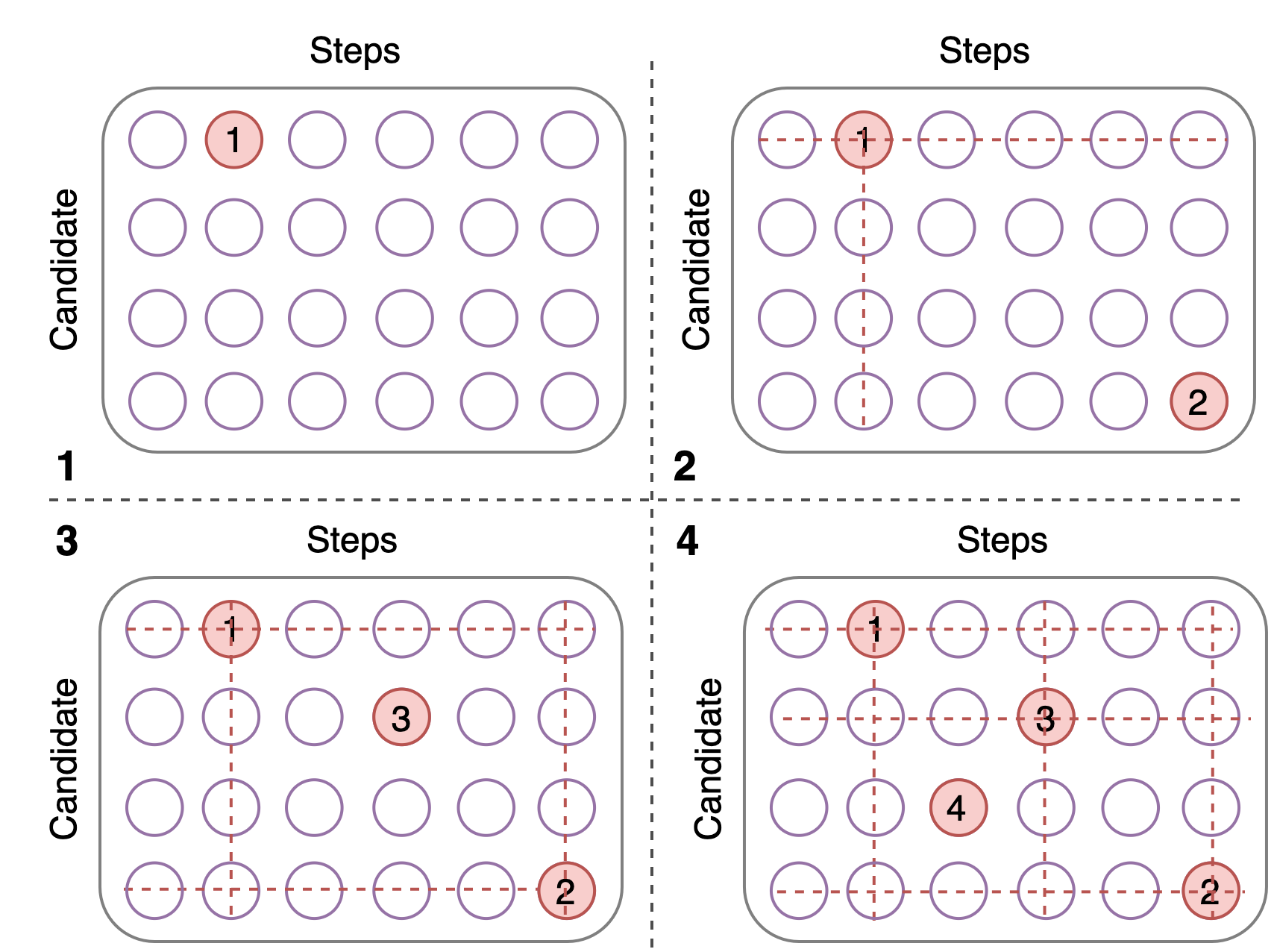}
    \caption{The matrix operation for constrained max-pooling}
    \label{fig:approach}
\end{figure}

For training the model, we define two different objectives directly applied to the similarity matrix. The textual cloze task does not have the direct supervision required for the alignment between candidates and steps, and our objective is designed to use the answer of the question to train this latent space of alignments. For imposing the constraint of the alignment to be disjoint between steps and candidates, one way is to simply compute the maximum of each row in the similarity matrix and use that as the aligned step for each candidate answer; However, we introduce constrained max-pooling which is a more sophisticated approach as shown in Figure~\ref{fig:approach}. We compare these two alternatives in the experimental results. We apply an iterative process to select the most related pair of instruction (a column) and answer candidate (a row) while removing the related column and row each time until all candidate answers find their aligned instruction. We denote the final selected maximum scores by $m = (S_{1i_{1}}, S_{2i_{2}}, S_{3i_{3}}, S_{4i_{4}})$, where $i_{c} \in [1, number\_of\_steps]$ is the index of the step with maximum alignment score with candidate $c$and for all pairs of candidates $c$ and $d$, $c \neq d \implies i_{c} \neq i_{d}$.

Respectively, we define two following objectives. The first objective maximizes the distance between the maximum score of the correct answer and the maximum score of another random wrong answer candidate. Furthermore, by fixing the instruction with the maximum alignment with the correct answer, it decreases the score of the other candidates alignments with that instruction. The second objective, increases the maximum similarity score of the answer to approach to 1 while decreasing the other maximum scores to be lower than $0.1$. 
\begin{align}
    \begin{split}
    \label{formula:one}
        Loss = \max(0, S_{ri_{r}} - S_{ai_{a}} + 0.1)+ \\
        \sum^{4}_{c \neq a}\max(0, S_{ci_{a}} - S_{ai_{a}} + 0.1) 
    \end{split} \\
    \begin{split}
    \label{formula:two}
        Loss = (1 - S_{ai_{a}}) + \sum^{4}_{c \neq a}\max(0, S_{ci_{c}} - 0.1)
    \end{split}
\end{align}

Where $a \in \{1, 2, 3, 4\}$ is the correct answer number and $r$ is a random index from $\{1, 2,, 3, 4\} - \{a\}$. The main difference in objective~\ref{formula:one} and objective~\ref{formula:two} is the regularization term on the selected instruction column in the alignment matrix.

\section{Experiment}
\subsection{Baselines}
\textbf{Hasty Student}~\cite{tapaswi2016movieqa} is a simple approach considering only the similarity between elements in question and candidate answers. This baseline fails to get good results due to the intrinsic of the task.

\noindent \textbf{Impatient Reader}~\cite{hermann2015teaching} computes attention from answers to the recipe for each candidate and despite being a complicated approach, yet it fails to get good results on the task. Moreover, multimodal Impatient reader approach uses both instructions and corresponding images.
\subsection{Results}
The RecipeQA textual cloze task contains $7837$ training, $961$ validation, and $963$ test examples. A learning rate of $4-e1$ is used for the first half and then $8-e2$ for the second half of training iterations. We use the momentum of $0.9$ for all variations of our model. We train for $30$ iterations with a batch size of 1 and optimize the weights using an SGD optimizer.
For word embedding, the pre-trained BERT embedding in Flair framework is used. For the image representations,  ResNet50~\cite{he2016deep} pre-trained on Imagenet~\cite{russakovsky2015imagenet} using PyTorch library~\cite{NEURIPS2019_9015} is applied.

Table~\ref{tab:results} presents the experimental results. We call the model variations which use the loss objective in Equation (\ref{formula:one}) as Model-obj~\ref{formula:one} and the ones that use the loss in Equation (\ref{formula:two}) as Model-obj~\ref{formula:one}. Using the objective in Formula (\ref{formula:one}) yields better results in all experiments. This indicates the benefit of using the column-wise disjoint constraint on the similarity matrix. Also, using multimodal information yields $1.12\%$ improvement. We elaborate further on the comparison between multimodal and unimodal results in Section \ref{sec:discussion}. 
\begin{table*}[h]
    \centering
    \begin{tabular}{lcc}
        \hline
         Models & Accuracy & p@2\\ \hline
         Human & 73.6 & - \\
         Hasty Student &  26.89 & - \\
         Impatient Reader & 28.03 & - \\
         Impatient Reader (multimodal) & 29.07 & - \\ \hline
         Model-Obj~\ref{formula:one} & 46.35 & \textbf{78.7}\\
         Model-Obj~\ref{formula:two} & 43.36 & - \\
         Model-Obj~\ref{formula:one} (multimodal) & 45.41 \\
         Model-Obj~\ref{formula:one} (multimodal) + LXMERT& \textbf{47.5} & 77.5\\ 
         Model-Obj~\ref{formula:one} (multimodal) + LXMERT - ConstrainedMaxPooling& 46.9 & 76.3 \\
         \hline
    \end{tabular}
    \caption{Evaluation on the test set}
    \label{tab:results}
\end{table*}

We provide our Pytorch implementation publicly available on Github~\footnote{https://github.com/HLR/LatentAlignmentProcedural}.
\subsection{Multimodal Results}
\label{sec:ablation}
In order to investigate the usefulness of the images in solving the textual cloze task, we propose two different models that incorporate the image representation in addition to the textual information of recipe steps. The first variation receives ResNet50 representations of the images and, after applying an LSTM layer, pulls the last layer as image representation. Finally, it concatenates the image representation to the question and instruction representation in the main architecture before applying the MLP and computing the cosine similarities. 

\begin{figure}[h]
    \centering
    \includegraphics[width=\linewidth]{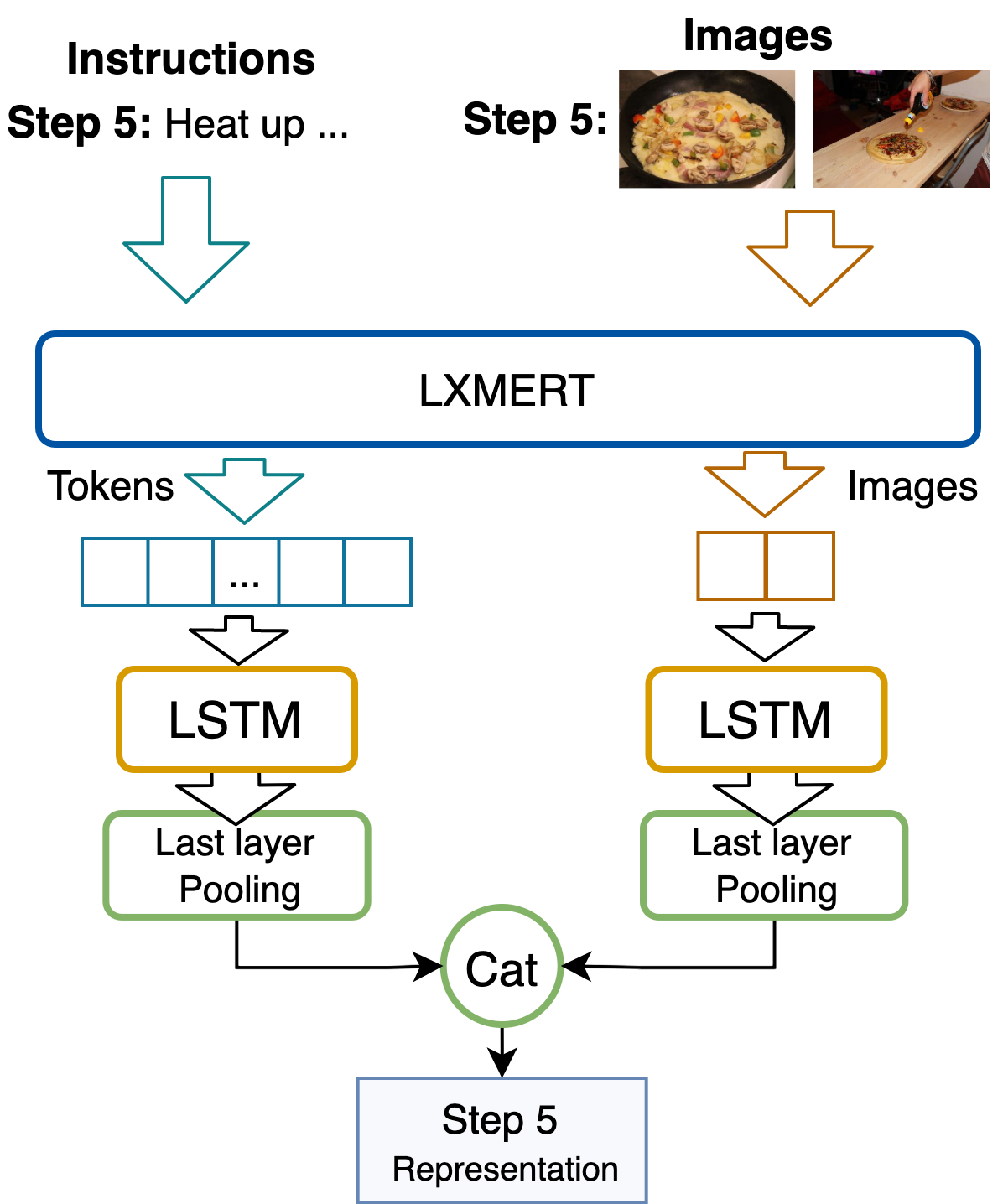}
    \caption{Using LXMERT for integrating multimodal information on steps}
    \label{fig:lxmert}
\end{figure}

The second variation as shown in figure \ref{fig:lxmert}, uses a more complex architecture introduced in LXMERT~\cite{tan-bansal-2019-lxmert}. We modify the architecture of LXMERT and apply it to the word embedding and image representations to flow the information from each to another. The updated word embedding and image representations are passed to an LSTM, and its last layer is used to represent the visual and textual information of a step. In the end, these representations are concatenated to each other and the question representation to build the instruction vector representation.
We report the results of these model variations in Table~\ref{tab:results}. Using the cross modality representations based on LXMERT provided extensive way to flow the information from text and image to each other and yields the best results.

\section{Discussion and Analysis}
\label{sec:discussion}
We did qualitative analysis using some examples and their results to better understand the behaviour of the proposed model.
Our model is almost able to detect all matched candidates with the instructions (in case that there exist multiple matches) but fails to choose the one that completes the sequence of the question items. 
This indicates the shortage of procedural hints inside our architecture while the latent alignment is proven to be practical.
By analysing the results, we found interesting cases where either multimodal or unimodal architectures could yield more accurate predictions.

\begin{figure}[h]
    \centering
    \includegraphics[width=\linewidth]{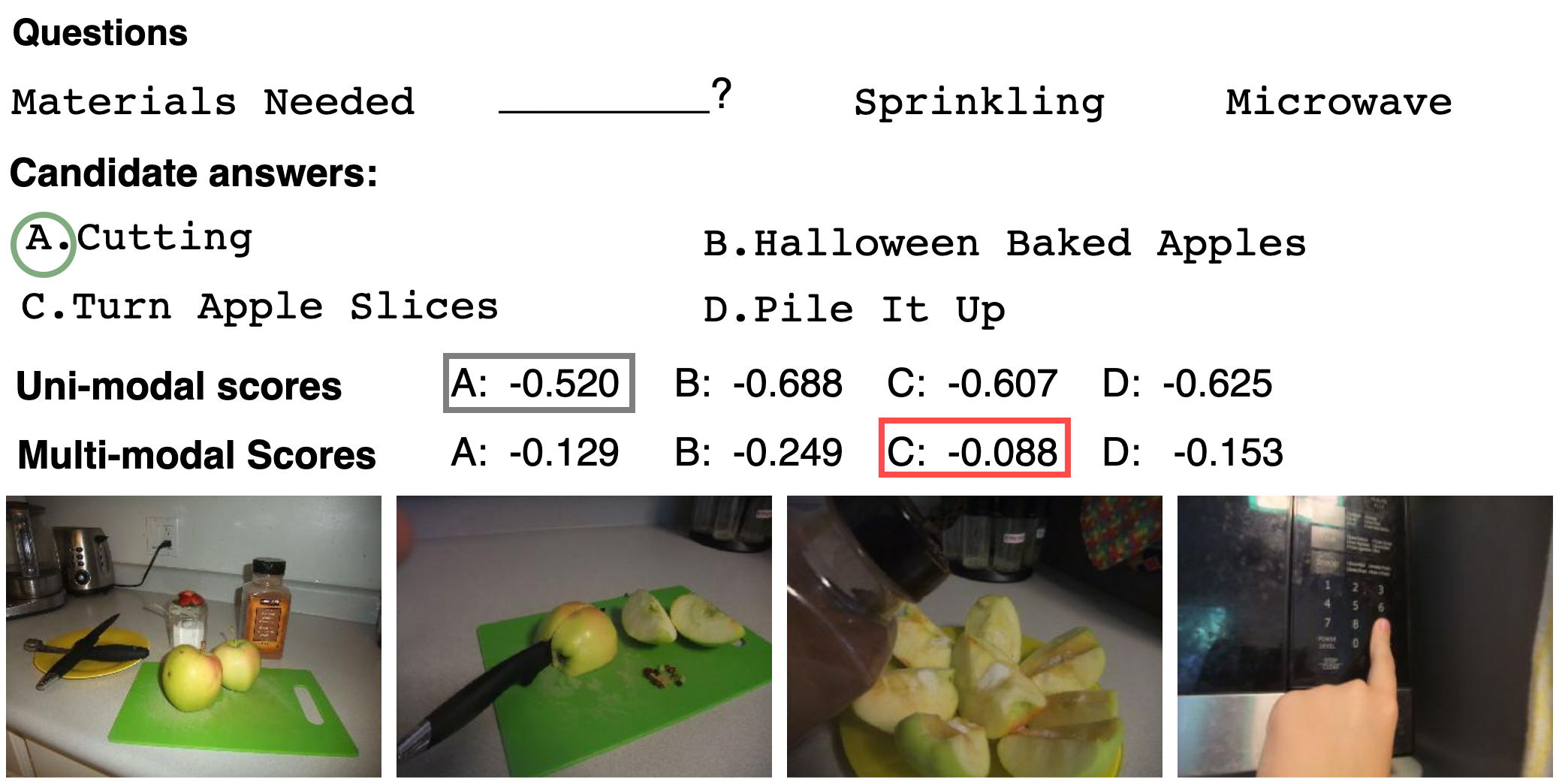}
    \caption{The image is misleading the multimodal setting to choose apple slices rather than cutting option}
    \label{fig:samplev-}
\end{figure}

\begin{figure}[h]
    \centering
    \includegraphics[width=\linewidth]{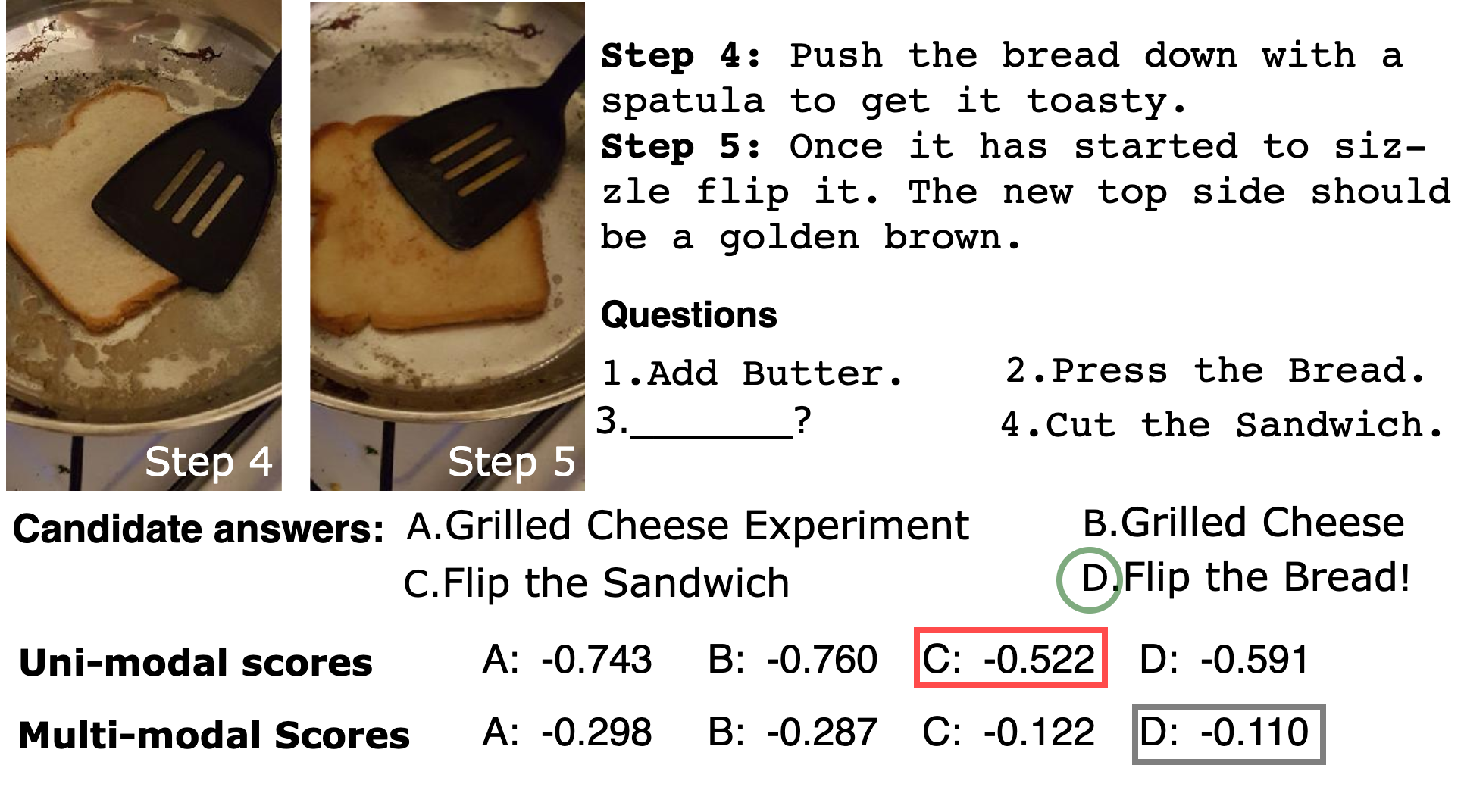}
    \caption{The images lead the model to understand that "it" refers to bread rather than sandwich}
    \label{fig:samplev+}
\end{figure}

\noindent\textbf{Multimodal - , Unimodal +}:
\begin{compactitem}
    \item Images contain misleading information~(see example in Figure \ref{fig:samplev-}).
    \item Image quality is low.
    \item Images are not showing the steps correctly.
    \item Text contains direct mentions of candidate answers.
\end{compactitem}
\noindent\textbf{Multimodal + , Unimodal -}:
\begin{compactitem}
    \item The sequence of the images provide detailed steps and good quality. 
    \item The entities in candidates answers are shown in the pictures but not in the text.
    \item The recipes instructions are very short and the images provide more information.
\end{compactitem}
In some cases, the multimodal information can fix the errors resulted from not considering the order of events in the proposed architecture. Our intuition is that, although, the textual model does not contain information from previous steps, the images carry useful information on what has been already done. An example of this is shown in Figure \ref{fig:samplev+}, where co-reference resolution is required to answer the question correctly.

Furthermore, we have tested our multimodal architecture with representations of ResNet101 and the results dropped. We confirmed this experiment by re-implementing Hasty Student approach on visual coherence task~(that has 68\% accuracy with ResNet50) and obtained 35\% lower than ResNet50. This can be due to the lack of quality of images resulting in extra noise when using a more complicated network. Thus, ResNet50 achieves better accuracy by producing more abstract representations of the images.

\section{Conclusion and Future Work}
We proposed a model for RecipeQA textual cloze task which exploits the latent alignment of question items with instructions. Moreover, we investigated the benefit of using multimodal information in this task by comparing three different architectures and provided qualitative analysis on some examples to justify the results. Our model exceeded the baselines and improved the SOTA by over 19\%.
As a future direction, we will investigate the usage of the latent alignment in other tasks. We will apply more complex methods on textual abstractions and attention mechanisms to link the candidate answers with the recipe instructions. Investigating how to incorporate the question order in the architecture is another direction.

\section*{Acknowledgments}
This work is (partially) supported by the Office of Naval Research grant N00014-19-1-2308.

\bibliography{main}
\bibliographystyle{acl_natbib}


\end{document}